\documentclass[journal]{IEEEtran}
\usepackage{pifont}

\usepackage{amsmath,amsfonts,amssymb}

\usepackage{algorithmic}
\usepackage{algorithm}
\usepackage{utfsym}
\usepackage{array}
\usepackage{multirow}
\usepackage{makecell}
\usepackage{booktabs}
\usepackage{tabulary}
\usepackage{cite}
\usepackage{graphicx}
\usepackage{colortbl}
\usepackage{newunicodechar}

\usepackage{hyperref}  
\usepackage{textcomp}  
\usepackage{stfloats}  
\usepackage{url}       
\usepackage{verbatim}  
\usepackage{enumitem}  
\usepackage{tikz}      
\usepackage{calc}      
\usepackage{xspace}    
\usepackage[caption=false,font=normalsize,labelfont=sf,textfont=sf]{subfig}

\def\ours{\textbf{BTDNet}\xspace}

\newcommand{\Rmnum}[1]{\expandafter\@slowromancap\romannumeral #1@}

\begin{document}

\title{Referring Remote Sensing Image Segmentation via Bidirectional Alignment Guided Joint Prediction}

\author{{Tianxiang Zhang, Zhaokun Wen, Bo Kong, Kecheng Liu, Yisi Zhang, Peixian Zhuang, Jiangyun Li}
\thanks{Zhaokun Wen, Bo Kong, Kecheng liu and Yisi zhang are with the School of Automation and Electrical Engineering, University of Science and Technology Beijing, Beijing 100083, China (e-mail: M202320772@xs.ustb.edu.cn, M202320748@xs.ustb.edu.cn, 18003506710@163.com, M202210579@xs.ustb.edu.cn).

Tianxiang Zhang, Peixian Zhuang and Jiangyun Li are with the Key Laboratory of Knowledge Automation for Industrial Processes, Ministry of Education, the School of Automation and Electrical Engineering, University of Science and Technology Beijing, Beijing 100083, China (e-mail: leejy@ustb.edu.cn, txzhang@ustb.edu.cn). Jiangyun Li is also with the Shunde Graduate School of University of Science and Technology Beijing, China. (Corresponding author: Jiangyun Li.)

This work was supported by the Natural Science Foundation of China under Grant 42201386, in part by Fundamental Research Funds for the Central Universities of USTB: FRF-TP-24-060A.}}

\markboth{IEEE TRANSACTIONS ON GEOSCIENCE AND REMOTE SENSING}%
{Shell \MakeLowercase{\textit{}}: A Sample Article Using IEEEtran.cls for IEEE Journals}


\maketitle

\begin{abstract}
Referring Remote Sensing Image Segmentation (RRSIS) is critical for ecological monitoring, urban planning, and disaster management, requiring precise segmentation of objects in remote sensing imagery guided by textual descriptions. This task is uniquely challenging due to the considerable vision-language gap, the high spatial resolution and broad coverage of remote sensing imagery with diverse categories and small targets, and the presence of clustered, unclear targets with blurred edges. To tackle these issues, we propose \ours, a novel framework designed to bridge the vision-language gap, enhance multi-scale feature interaction, and improve fine-grained object differentiation.
Specifically, \ours introduces: (1) the Bidirectional Spatial Correlation (BSC) for improved vision-language feature alignment, (2) the Target-Background TwinStream Decoder (T-BTD) for precise distinction between targets and non-targets, and (3) the Dual-Modal Object Learning Strategy (D-MOLS) for robust multimodal feature reconstruction. Extensive experiments on the benchmark datasets RefSegRS and RRSIS-D demonstrate that \ours achieves state-of-the-art performance. Specifically, \ours improves the overall IoU (oIoU) by 3.76 percentage points (80.57) and 1.44 percentage points (79.23) on the two datasets, respectively. Additionally, it outperforms previous methods in the mean IoU (mIoU) by 5.37 percentage points (67.95) and 1.84 percentage points (66.04), effectively addressing the core challenges of RRSIS with enhanced precision and robustness. Datasets and codes are available at \url{https://github.com/wzk913ysq/BAJP}.
\end{abstract}


\section{Introduction}

\IEEEPARstart{R}{efering} Remote Sensing Image Segmentation (RRSIS)\cite{yuan2024rrsis,liu2024rotated} has emerged as a critical research focus in remote sensing image analysis \cite{cheng2016survey}, aiming to segment instance targets for remote sensing (RS) images based on textual descriptions. Unlike traditional domain-specific remote sensing image segmentation \cite{kotaridis2021remote,dong2023distilling} that  are constrained by a limited number of semantic labels, the RRSIS task enables open-domain segmentation by utilizing free-form textual descriptions as semantic referring information. This approach provides numerous applications in areas such as text-guided environmental monitoring\cite{he2016deep}, land cover classification\cite{talukdar2020land}, precision agriculture\cite{weiss2020remote}, and urban planning\cite{yan2015urban}, where specific objects or regions need to be identified and segmented based on natural language descriptions. By providing more flexibility in interpreting and processing remote sensing data, RRSIS enhances the ability to extract detailed, context-specific information from complex RS images.

The rapid advancements in multimodal representation learning \cite{guo2019deep} have significantly elevated the importance of RRSIS by effectively integrating visual and textual information. These multimodal methods enhance the accuracy and robustness of segmentation tasks, particularly in complex remote sensing environments. Such ability to align and fuse information across modalities is critical for addressing the challenges inherent in RRSIS, such as handling ambiguous targets and fine-grained object differentiation. As RRSIS relies on both image content and textual descriptions, these multimodal techniques are critical for improving the performance and applicability of RRSIS in real-world remote sensing applications. 

RRSIS tackles the challenge of identifying instance-level targets within the same category distinguished by various attributes, laying the foundation for precise context-aware target segmentation. However, several challenges differentiate RRSIS from natural image referring tasks. The first challenge is lying in the substantial gap between vision and language in remote sensing during the referring stage. Unlike natural images, which exhibit inherent correlations with natural language descriptions and enable dense relationships effectively leveraged in benchmark models (e.g., CLIP\cite{radford2021learning}) for visual understanding, remote sensing lacks models with strong vision-language alignment. Consequently, this gap leads to significant discrepancies between vision and text feature distributions, making cross-modal alignment in the RS domain substantially more challenging.

Due to the significant vision-language gap in remote sensing, two critical issues arise, severely affecting segmentation accuracy. First, the wide coverage of satellite/UAV-based remote sensing imagery leads to diverse object categories, with targets appearing at different scales. This variation, particularly for small objects, poses substantial challenges for precise localization and segmentation. The lack of robust vision-language alignment further weakens the model’s ability to distinguish between different categories, making fine-grained object recognition more difficult.
Second, the segmentation challenge is further exacerbated when objects of the same category are clustered together or when different objects are closely connected, making boundary delineation particularly difficult. Many ambiguous targets exhibit blurred edges, increasing the likelihood of mis-segmentation. Without a well-aligned vision-language model, distinguishing object boundaries based solely on textual descriptions remains difficult, further constraining segmentation accuracy in RRSIS.
\begin{figure}[!t]
    \centering
    \includegraphics[width=1\linewidth]{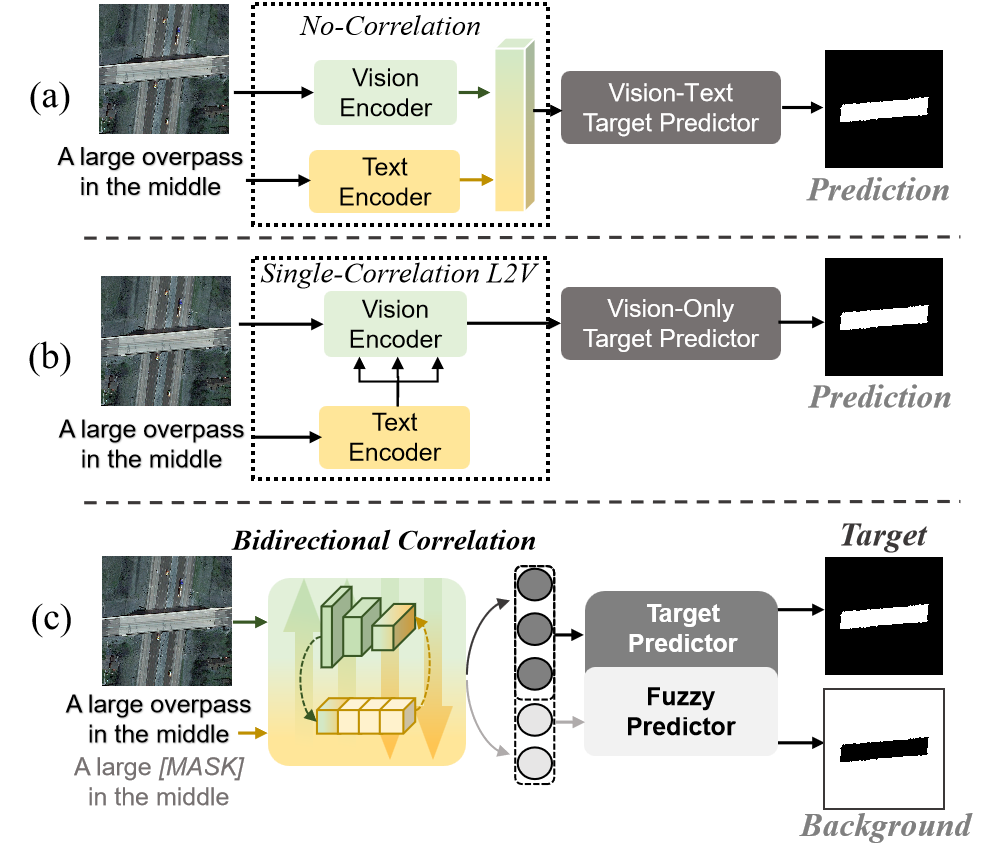}
    \vspace{-8mm}
    \caption{The illustration of the pipeline comparison between existing methods and our design for the RRSIS task.}
    \label{fig:intro}
\end{figure}


A lot of significant work has been made in this domain, most existing RRSIS algorithms~\cite{yuan2024rrsis,liu2024rotated} are adaptations of referring image segmentation (RIS) methods originally designed for natural images. As illustrated in Fig. \ref{fig:intro} (a) and (b), these approaches predominantly follow two schemes. The first scheme adopts a straightforward no-correlation pipeline, starting with independent modal encoding followed by cross-modal decoding\cite{ye2019cross,wang2022cris}. This approach involves high-dimensional cross-modal feature fusion, where features across modalities are roughly aligned. The mask prediction is jointly achieved by integrating both textual and visual features. The second scheme employs a more interactive pipeline\cite{yuan2024rrsis, liu2024rotated}. Textual information is incorporated into the vision encoding process to guide feature extraction, aligning visual features gradually from low to high dimensions. Unlike the aforementioned no-correlation method, this scheme involves multiple rounds of cross-modal correlation during the encoding and enhancement of visual information, enabling the decoding phase to rely on image features for mask prediction. While these methods achieve specific results, they face significant limitations in addressing the vision-language distribution gap inherent in RS imagery. This limitation becomes especially pronounced in RS scenes with numerous similar regions, scattered small targets, and unclear distinctions between foreground and background. Based on the unique characteristics of remote sensing images, we conducted targeted research and design. Our primary focus is on minimizing the vision-language distribution gap in remote sensing images by enhancing image-text interaction and information interpretation. Additionally, we designed refined localization strategies for small and ambiguous targets to address the challenges posed by complex object categories and high intra-class similarity.


To cope with the aforementioned challenges, we propose a novel method shown in Fig. \ref{fig:intro} (c) \ours, which performs \textbf{B}idirectional feature \textbf{A}lignment between vision and language. Furthermore, by leveraging auxiliary inputs masking key textual reference target information, this method enables \textbf{J}oint \textbf{P}redictions of foreground objects and background information, leading to more accurate RRSIS segmentation results. In particular, our proposed Bidirectional Spatial Correlation Module (\textbf{BSC}) is integrated into the bidirectional feature extractor to enable effective bidirectional feature interaction, bridging the gap between vision and language modalities. To handle the ambiguous characteristic of RS imagery, we designed the Target-Background TwinStream Decoder (\textbf{T-BTD}), using masked relational text as prior knowledge to differentiate between target and non-target entities. This strategy facilitates a comprehensive scene understanding of foreground objects and background regions by joint predictions. In addition, the Dual-Modal Object Learning Strategy (\textbf{D-MOLS}) is introduced addressing the vision-language gap prevalent in the remote sensing domain. By reconstructing critical textual information, D-MOLS enhances the alignment of features across modalities, thereby improving the understanding of language information in the RS scene. Extensive experiments conducted on two widely used benchmarks, RefSegRS~\cite{yuan2024rrsis} and RRSIS-D~\cite{liu2024rotated}, demonstrate that our proposed method, \ours, achieves state-of-the-art (SOTA) performance in the RRSIS task, particularly in high-precision and fine-grained segmentation. Qualitative comparisons through visualizations further highlight the robust performance of \ours across various remote sensing scenarios and for different segmentation targets.
In summary, the main contributions of this work are as follows:  
\begin{enumerate}
    \item We proposed a novel framework named \ours for the RRSIS task, focusing on bidirectional multimodal feature interaction and fore-background target joint prediction. Our method obtains state-of-the-art results on two public remote sensing datasets.
    \item Our proposed Bidirectional Spatial Correlation Module and dual-modal object learning strategy effectively bridging the vision-language distribution gap in the remote sensing domain, enhancing the model's multimodal alignment capability.
    \item By introducing the Target-Background TwinStream decoder into the design, our method effectively addresses the challenges posed by discrete and ambiguous targets prevalent in remote sensing scenarios.
\end{enumerate}

\section{RELATED WORK}
\subsection{Referring Image Segmentation for Natural Images}
Referring Image Segmentation (RIS), an extension of visual grounding \cite{yang2019fast,deng2021transvg}, aims to segment specific regions in an image based on textual descriptions. This task requires not only precise alignment of text and image features but also the ability to accurately delineate object boundaries, handle fine-grained details, and manage object appearance variations across different scales and contexts.

In the early stages, RIS models primarily used convolutional and recurrent neural networks to extract features, which were then combined for joint prediction. Hu et al. first introduced the RIS task to address challenges in existing semantic segmentation tasks when dealing with complex referential texts\cite{hu2016segmentation}. Since then, Li \textit{et al}. and Nagaraja \textit{et al}. employed Convolutional Neural Networks (CNNs) and Recurrent Neural Networks (RNNs) for processing bimodal information and performing joint mask prediction\cite{li2018rrn,nagaraja2016modeling}. By incorporating a modal interaction structure, Liu \textit{et al}. improved cross-modal alignment\cite{liu2017recurrent}. Moreover, Margffoy \textit{et al}. introduced a dynamic multimodal network to integrate recursive visual and linguistic features\cite{margffoy2018dynamic}.

Follow-up studies emphasize the importance of effective cross-modal interactions, with self-attention mechanisms serving as a cornerstone for facilitating such interactions. For example, Ye \textit{et al}. proposed a cross-modal self-attention module to capture long-range dependencies, paired with a gated multi-level fusion module for feature integration\cite{ye2019cross}. Similarly, Hu \textit{et al}. introduced a bidirectional cross-modal attention module to enhance alignment between linguistic and visual features\cite{hu2020bi}. Moreover, Shi \textit{et al}. developed a keyword-aware network that leverages keywords to refine region relationships\cite{shi2018key}.

With the continuous evolution of deep learning architectures, Transformers have emerged as the leading approach for RIS, offering superior global modeling and cross-modal alignment. Transformer-based methods, such as CGAN\cite{luo2020cascade}, LAVT\cite{yang2022lavt}, RESTR\cite{kim2022restr}, M3Att\cite{liu2023multi}, MagNet\cite{chng2024mask}, have significantly advanced the field. Kim \textit{et al}. introduced the first convolution-free architecture, leveraging Transformers for long-range interactions\cite{kim2022restr}. Liu \textit{et al}. proposed multi-modal mutual attention and a mutual decoder for improved feature integration\cite{liu2023multi}. Luo \textit{et al}. designed a cascaded group attention network for iterative reasoning\cite{luo2020cascade}, while Chng \textit{et al}. introduced a text reconstruction task for fine-grained cross-modal alignment\cite{chng2024mask}.

In conclusion, while RIS research has progressed from traditional to Transformer-based architectures, achieving notable improvements in multimodal interaction and alignment, it continues to face challenges such as fine-grained object differentiation, managing occlusions, and accurately aligning spatially distributed objects with their corresponding textual descriptions. Additionally, context-dependent understanding and the inherent ambiguity of language remain significant hurdles.

\subsection{Referring Remote Sensing Image Segmentation}
Identifying objects in remote sensing images typically requires specialized expertise. Referring Remote Sensing Image Segmentation (RRSIS) has gained attention as it helps non-expert users extract precise information based on textual descriptions. This task is closely related to visual grounding in remote sensing \cite{sun2022visual}, where textual cues are aligned with specific image regions. However, research in this area is still in its early stages\cite{yuan2024rrsis,liu2024rotated} with limited studies available. 

Liu \textit{et al}. introduced the RRSIS task, specifically designed to segment targets in remote sensing images based on natural language expressions\cite{liu2024rotated}. This work references the LAVT model proposed by\cite{yang2022lavt} for the natural image referring segmentation task. To address the common occurrence of small and dispersed objects in remote sensing images, \cite{yuan2024rrsis} designed the cross-scale enhancement module that effectively utilizes language guidance to integrate deep and shallow visual features, enhancing the model's ability to discern small targets. Liu \textit{et al}. proposed a model named RMSIN that deal with the complex spatial scale and directional issues in remote sensing images through rotated convolutions\cite{liu2024rotated}.
\begin{figure}
    \centering
    \includegraphics[width=1\linewidth]{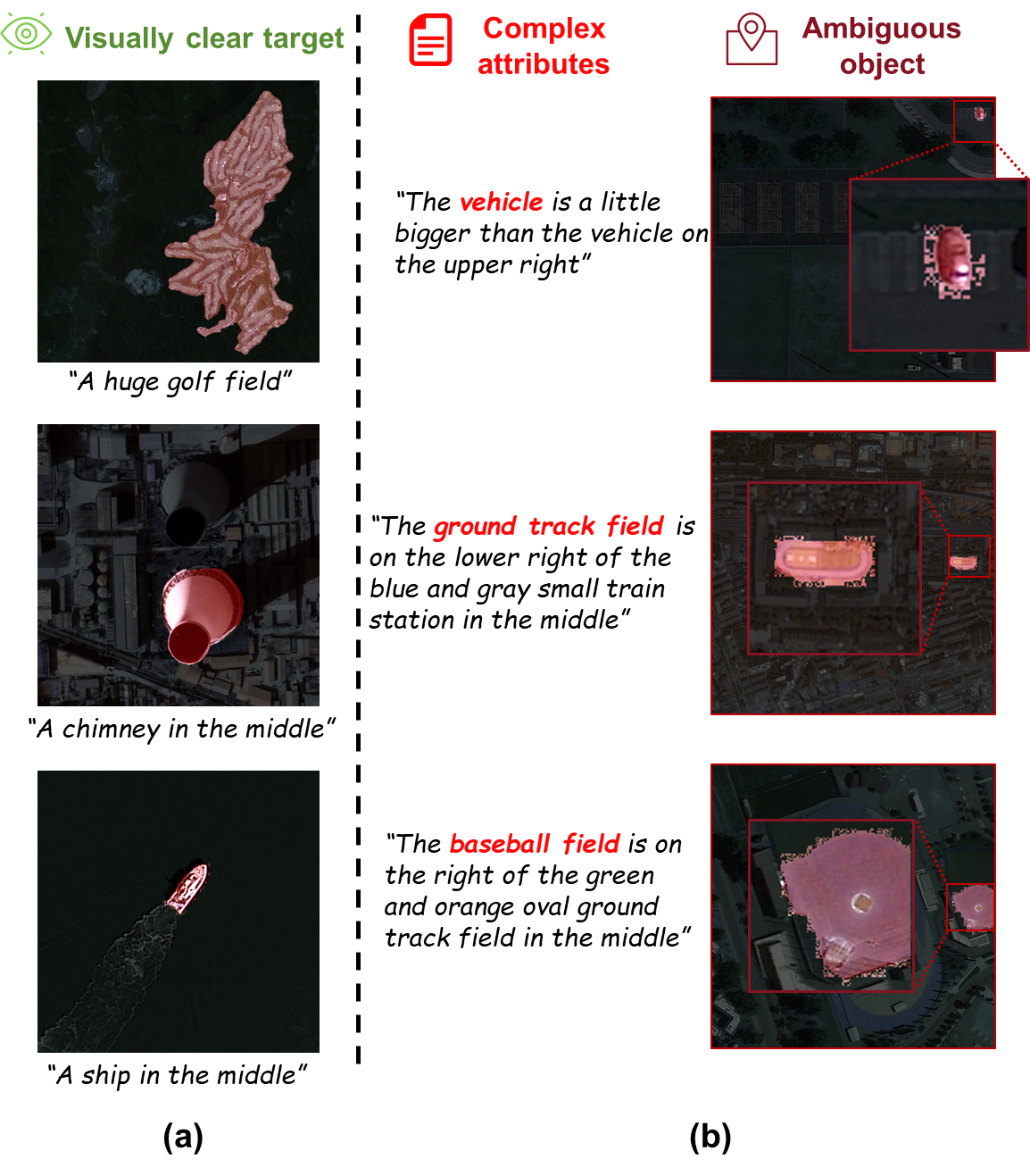}
    \vspace{-5mm}
    \caption{A comparison of two scenarios. (a) shows clear targets with simple descriptions, demonstrating the superior performance of existing methods. (b) illustrates complex descriptions with ambiguous backgrounds, highlighting the limitations of current approaches. }
    \label{fig:relate}
    
\end{figure}

In general, as shown in Figure \ref{fig:relate}, existing methods for referring remote sensing image segmentation achieve good performance when coping with visually clear objects and brief descriptions. However, their effectiveness diminishes when faced with complex textual relationships, as their capabilities in textual feature extraction and visual-textual modality alignment are limited, leading to inferior results. Additionally, these models cannot effectively address potential confusion between targets and backgrounds in remote sensing images.
\begin{figure*}[!t]
    \centering
    \includegraphics[width=1\linewidth]{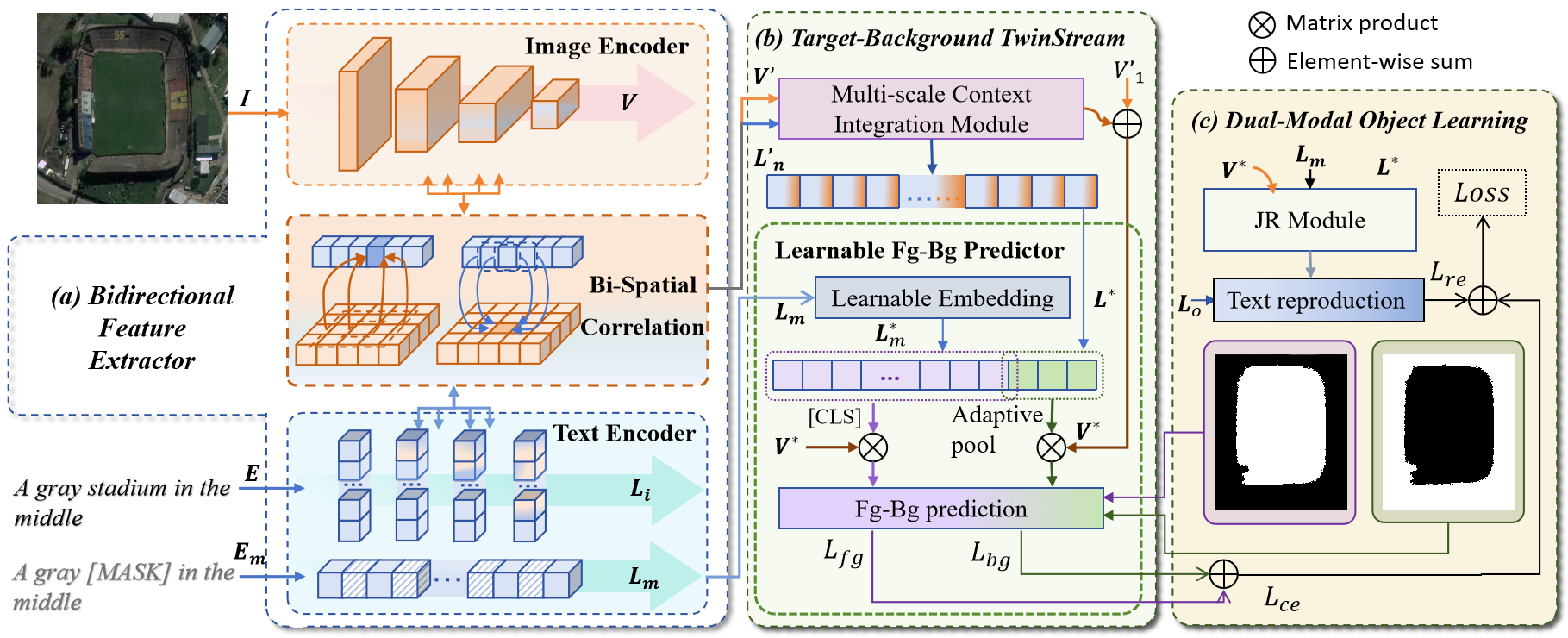}
    \vspace{-6mm}
    \caption{The overall framework of the proposed \ours Network includes the following components: \textbf{(a)} \textbf{Bidirectional Feature Extractor}, in which the visual and text encoders extract features from image and text inputs, respectively, with the Bidirectional Spatial Correlation Module enabling bidirectional information exchange at the feature level. \textbf{(b)} \textbf{Target-Background TwinStream}, implementing a text-aware dual-stream inference strategy for entity targets and category-agnostic areas during the mask prediction stage. \textbf{(c)} \textbf{Dual-Modal Object Learning}, designed to apply a reconstruction loss named \textit{Lre} after erasing text entity words, in which the reconstruction module leverages multimodal visual features to guide text reconstruction.}
    \vspace{-3mm}
    \label{fig:method}
\end{figure*}
\section{Method}



\subsection{Problem Formulation}

The primary goal of this study is to address the task of referring image segmentation in remote sensing (RS) data, where the objective is to generate precise masks for target objects based on natural language expressions. Formally, let $I \in \mathbb{R}^{C\times H \times W }$ represent an RS image, where $H, W, C$ denote the height, width, and number of channels, respectively. A corresponding textual description $E = {e_1, e_2, \dots, e_n}$ provides semantic guidance for the segmentation task, with $n$ being the number of tokens in the expression.

The task is to produce an output segmentation mask $O \in {0, 1}^{H \times W}$ such that each pixel $p \in I$ is assigned a binary label indicating whether it belongs to the region referred to by $E$. Given a dataset $\Omega = {(I_i, E_i, G_i)}_{i=1}^{Num}$, where $G_i \in {0, 1}^{H \times W}$ is the ground truth mask and $Num$ is the number of samples, the aim is to design a model $f$ that maps $(I, E)$ to $O$ by learning the alignment between textual and visual features. This study focuses on evaluating and enhancing the multi-scale and cross-modal understanding of RS images, particularly in handling the challenges posed by high-resolution data, ambiguous object attributes, and fine-grained target localization.

\subsection{Overview Architecture}
Fig. \ref{fig:method} presents the overall architecture of our framework, referred to as \ours. It comprises three main components: the Bidirectional Feature Extractor, the Target-Background TwinStream Decoder, and the Dual-Modal Object Learning Strategy. 

Given an image-text pair \((I, E)\) as input, the textual component first undergoes masking of key subject phrases, resulting in the modified text \(E_m\). The Bidirectional Feature Extractor performs hierarchical feature analysis and interaction on \((I, E)\). Within this component, the visual encoder and text encoder perform multi-stage feature sampling and analysis. At each stage, the Bidirectional Spatial Correlation module facilitates spatial-level feature interaction, enabling bidirectional information exchange. This process generates vision-oriented multimodal features \(V'\) and text-oriented multimodal features \(L'_n\). Meanwhile, the masked text \(E_m\) is encoded to produce the textual features \(L_m\).

In the Target-Background TwinStream decoder, \(V'\) and \(L'_n\) are integrated through a Multi-scale Context Integration module, resulting in multi-scale contextual attention features \(L^*\) and \(V^*\) for pixel-level mask prediction. Meanwhile, the referential feature \(L_m\) is embedded into learnable representations, yielding \(L^*_m\). The foreground and background indicators \(L^*\) and \(L^*_m\) are then combined with \(V^*\), enabling the model to predict masks for referred targets and category-agnostic objects.

Additionally, our Rephrase Module leverages vision-oriented multimodal features to assist in reconstructing the masked text \(E_m\), enhancing the semantic understanding of the model. The subsequent sections provide a detailed explanation of each component's construction.

\subsection{Bidirectional Feature Extractor}
Given an image-text pair \( I \in \mathbb{R}^{3 \times H \times W} \) and \( E \in \mathbb{R}^{D \times N} \), where \( E \) is the text embedding from a tokenizer, we first use the NLTK library~\cite{bird2006nltk} to mask key objects in the text description, replacing them with padding. This results in a modified language expression, \( E_m \in \mathbb{R}^{D \times N} \), which is then processed by the BERT encoder~\cite{devlin2018bert} to obtain category-agnostic textual features \( L_m \in \mathbb{R}^{D \times N} \). Simultaneously, the BERT encoder is partitioned into stages at layers 3, 6, 9, 12 to align with hierarchical features from the vision branch, enabling bidirectional interaction during feature extraction.

The complete image-text pair \((I, E)\) is jointly fed into the multi-stage feature encoder. At each stage, a \textit{Bidirectional Spatial Correlation Module} injects information from one modality into another. We perform an "Unfold" operation on the encoded image-text features to obtain local features with multiple receptive fields. These features reweight the object modality, enabling the extractor to focus on fine-grained local contextual information. Through progressive interaction, this module generates vision-oriented and text-oriented multimodal features: \(V' = \{ V'_i \mid V'_i \in \mathbb{R}^{C_i \times H_i \times W_i}, \, i = 1, 2, 3, 4 \} \) and \(L' = \{ L'_i \mid L'_i \in \mathbb{R}^{D \times N}, \, i = 1, 2, 3, 4 \}\) This multi-stage process enables progressively more profound cross-modal information fusion.

\subsubsection{Bidirectional structure}
The feature extraction backbones for the vision and text branches adopt Swin Transformer~\cite{liu2021swin} and BERT~\cite{devlin2018bert}, respectively. These backbones are divided into multi-stage feature encoders to perform multi-scale and deep information extraction across four stages while enabling bidirectional cross-modal interactions at each stage.

The input image \( I \) and text \( E \) are encoded sequentially at each stage, generating visual features \( V_i \) and text features \( L_i \) as inputs for modality interaction. The Bidirectional Spatial Correlation module fuses \( V_i \) and \( L_i \), producing cross-modality enhanced features \( V_i' \) and \( L_i' \), thereby completing the $i$-th stage of feature extraction and interaction. These enriched features are then reintroduced into the respective encoders for subsequent stages.

For each stage \( i = 1, 2, 3, 4 \), the enhanced features are computed as:
\begin{equation}
\label{feature}
V_i', L_i' = \text{BSC}(V_i, L_i) + (V_i, L_i)
\end{equation}
These features are passed to the next stage's encoder to obtain \( V_{i+1} \) and \( L_{i+1} \), while the masked text \( E_m \) is processed by BERT to extract \( L_m \).
\subsubsection{Bidirectional Spatial Correlation}
\begin{figure}[!t]
\centering
\includegraphics[width=2.8in]{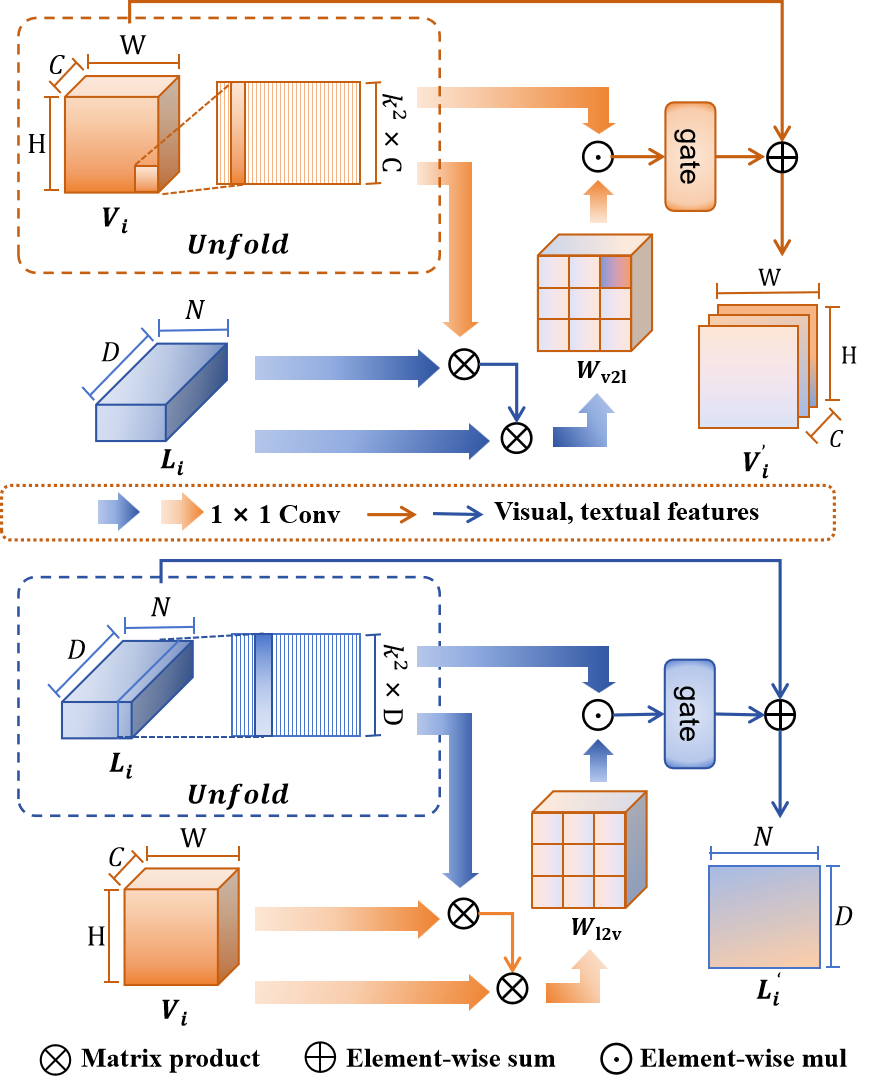}
\vspace{-2mm}
\caption{The illustration of Bidirectional Spatial Correlation Module. The correlation of information between image-to-text and text-to-image will be performed based on spatial context.}
\label{fig:details}
\end{figure}
As illustrated in Fig. \ref{fig:details}, for each stage i, the BSC module takes the visual features \( V_i \) and the corresponding textual features \( L_i \) as input. We design a bidirectional spatial correlation mechanism that leverages rich intra-modal context and cross-modal interactions to integrate visual and textual features across stages effectively. Specifically, the fusion process comprises vision-oriented and text-oriented feature fusion components. The corresponding fusion procedure is detailed as follows.
For each stage \( i \), the input visual and textual unimodal features \( V_i \) and \( L_i \) are cropped along the spatial dimension to extract per-pixel and per-token contextual information, the specific process as follows:
\begin{equation}
\begin{cases}
v_i^{(h,w)} = \text{Unfold}(V_i, k) \in \mathbb{R}^{ M }, \\
l_i^{len} = \text{Unfold}(L_i, k) \in \mathbb{R}^{ M }.
\end{cases}
\label{eq:contextual_features}
\end{equation}
Here, \( k \) represents the size of the contextual receptive field window. In the visual and textual branches, the embedding dimensions are defined as \( k^2 D \) and \( k^2 C_i\), respectively. \( h \), \( w \) and \( len \) correspond to spatial indices for pixels and word tokens. The $\text{Unfold}(·)$ operation slides a \( k \times k \) window across the feature map, extracting local patches centered around each pixel or word token. This produces feature tensors, where \( v^{(h,w)} \) and \( l^{len} \) represent pixel-wise and token-wise contextual information, respectively, under the receptive field of size \( k \). We denote the set of local vectors as \( W_v = \{ v^{(h, w)} \}\in \mathbb{R}^{H_iW_i \times M} \) and \( W_l = \{ l^{len} \} \in \mathbb{R}^{D\times M}\).

We then perform an initial mapping at the local feature set $W_v$ and $W_l$ to calculate the fine-grained visual-textual bidirectional affinity weight matrices \( W_{\text{v2l}} \in \mathbb{R}^{H_iW_i \times D} \) and \( W_{\text{l2v}} \in \mathbb{R}^{D \times H_iW_i} \), which are weighted according to different receptive fields \( k \). The specific formula is described as follows:


\begin{equation}
\begin{cases}
W_{\text{v2l}} = \sum\limits_{k \in \{1, 3, 5\}} \left\{ \alpha_k (W_v \otimes L_i) \right\} , \\
W_{\text{l2v}} = \sum\limits_{k \in \{1, 2, 3\}} \left\{ \beta_k (W_l \otimes V_i^f) \right\} .
\end{cases}
\label{eq:W_affinity}
\end{equation}

Here, \( \alpha_k \) and \( \beta_k \) represent learnable weighting parameters that reflect the importance of sampling with different receptive fields. The symbol \( \otimes\) denotes matrix multiplication. \( V_i^f \in \mathbb{R}^{C_i \times W_iH_i} \) represents the flattened visual features \( V_i \), where the spatial dimensions \( H \) and \( W \) are unfolded into a single dimension. 

Based on the affinity weight matrices \( W_{\text{v2l}} \) and \( W_{\text{l2v}} \), we reweight the visual and textual features using the cross-modal information to match the domain differences between the two modalities. Finally, a gating unit concatenates the cross-aligned multimodal features, which are then returned to the feature extraction backbone through a residual connection for bidirectional correlation, yielding \( V_i' \) and \( L_i' \), which will be used in deeper feature extraction in the subsequent stage. The process is formulated as follows:
\begin{equation}
\begin{cases}
V_i' = \text{Gate}_v \left\{ W_{\text{v2l}} \otimes L_i \right\} \circledcirc V_i + V_i, \\
L_i' = \text{Gate}_l \left\{ W_{\text{l2v}} \otimes V_i^f \right\} \circledcirc L_i + L_i.
\end{cases}
\label{eq:cross_modal_fusion}
\end{equation}
Here, $\text{Gate}(\cdot)$ represents a gating unit composed of a $1 \times 1$ convolution followed by an InstanceNorm layer, which dynamically controls the importance of the features. The symbol $\otimes$ denotes matrix multiplication, while $\circledcirc
$ represents element-wise multiplication.

\subsection{Target-Background TwinStream}
\begin{figure}[!t]
    \centering
    \includegraphics[width=2.65in]{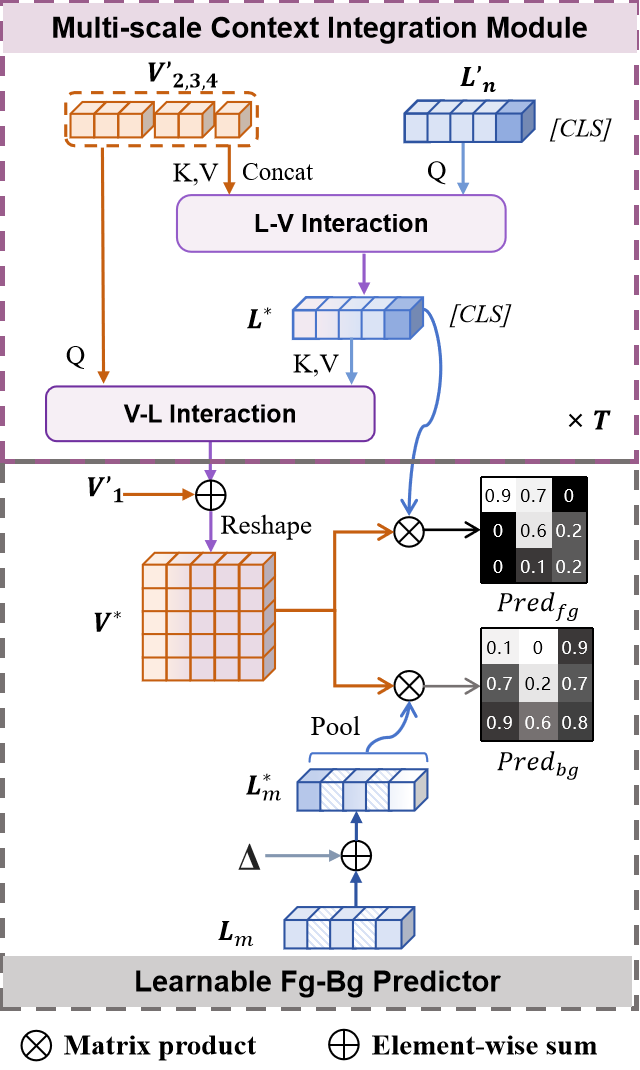}
    \vspace{-2mm}
    \caption{The illustration of the Target-Background TwinStream Decoder, comprising the Multi-scale Context Integration for multi-scale feature enhancement and the Learnable Fg-Bg Predictor for distinguishing target and background regions. }
    \label{fig:decoder}
\end{figure}

Fig. \ref{fig:decoder} illustrates the decoding process of the Target-Background TwinStream Decoder (T-BTD). T-BTD enhances multimodal features \( V' \) and \( L'_n \) via the Multi-scale Context Integration module, achieving cross-modal alignment to produce \( L^* \) and \( V^* \) for predicting target and category-agnostic masks. Masked semantic features \( L_m \) provide prior knowledge for peripheral information.

In the Learnable Fg-Bg Predictor, \( L_m \) is enriched through embedding, generating \( L_m^* \). \( L^* \) and \( L_m^* \) are matched with visual features to predict pixel-level foreground (\( O_{\text{fg}} \)) and background (\( O_{\text{bg}} \)) masks, with cross-entropy loss computed against ground truths.


\subsubsection{Multi-scale Context Integration Module}
The Multi-scale Context Integration (MCI) module captures the bidirectional interaction in this work, addressing the limitations of multi-scale and textual context associations in dual-modality features extracted during the feature extraction phase. To achieve robust semantic alignment across resolutions, we first reduce the channel dimension \( C_i \) of the input feature \( V'_i \) to obtain\( V_{\text{con}} \in \mathbb{R}^{D \times S} \): 
\begin{equation}
\label{deqn_ex1dda}
V_{\text{con}} = \text{Flat} \left( \text{Concat} \left( \left\{ \phi(V'_i) \right\}_{i=2}^4 \right) \right)
\end{equation}
Where $\phi(·)$ applies a $1 \times 1$ convolution to reduce the channel dimension of \( V'_i \), $\text{Concat}(·)$ integrates the features extracted from stages \( i=2, 3, 4 \) along the spatial dimension, leveraging multi-level semantic representations from deeper network layers, and $\text{Flat}(·)$ reshapes the spatial dimensions of the tensor ($H_i \times W_i$) into a single dimension \( S \). Here, \( S = \sum_{i=2}^4 H_i W_i \) represents the total spatial locations across stages.

In the MCI, Cross-Attention operations are performed between \( V_{\text{con}} \) and \( L'_n \), facilitating the alignment of textual descriptions with image features. This addresses the challenge of insufficient context in free-form descriptions of remote sensing images. The V-L interaction can be formulated as:
\begin{equation}
\label{MSAtFDFtn}
L^* = \text{Proj}(\text{FFN}(\text{CrossAtt}(L'_n, V_{\text{con}})))
\end{equation}
Here, \( L'_n \) serves as the query in the Cross-Attention mechanism, while \( V_{\text{con}} \) functions as both the key and value. FFN and Proj represent the feed-forward network and projection network, respectively.

Subsequently, the updated \( L^* \in \mathbb{R}^{D \times N} \), which integrates multi-scale prior information with visual localization, guides pixel-level textual matching in the visual branch. This L-V interaction process can be formulated as follows:
\begin{equation}
\label{MSAttn}
V^* = \text{Proj}(\text{MSAttn}(\text{CrossAtt}(V_{\text{con}}, L^*))).
\end{equation}
Here, \( V_{\text{con}} \) is used as the query, and \( L^* \) serves as both the key and value. Additionally, we employ multi-scale attention to facilitate multi-scale interactions. Finally, we obtain context-enhanced multi-scale visual features \( V^*\in \mathbb{R}^{D \times S} \). 
At this stage, \( V^* \) and \( L^* \) will be returned as the input of MCI for multiple interactions. Ultimately, this module is executed for \( T \) times.
\subsubsection{Learnable Fg-Bg Predictor}
After obtaining spatial-aware \( L^* \) and semantic-aware $V^*$, we reshape $V^*$ to \( V_{\text{attn}} = \{ V_{\text{attn}} \mid V^i_{\text{attn}} \in \mathbb{R}^{H_i \times W_i \times D}, i =  2, 3, 4 \} \). Then, \( V^2_{\text{attn}} \) is upsampled to \( H_1 \times W_1 \) via bilinear interpolation and added to the \( V'_1 \) feature to obtain \( V_{\text{pred}} \in \mathbb{R}^{H_1 \times W_1 \times D} \), which is used for mask prediction. To address ambiguity in remote sensing images, where unmentioned objects cause segmentation issues, we introduce a background prediction branch using masked textual features \( L_m \) as prior information. These features are passed through a learnable embedding and combined with learnable parameters as follows:
\begin{equation}
\label{deqn_ex241a}
L^*_m = \text{Concat}\left( \{\text{AP}_{r_i}(L_m)\}_{i=1}^R \right) +\Delta
\end{equation}
Here, \( \Delta \) denotes the learnable parameter matrix, and \( \text{AP}_{r_i} \) denotes the adaptive pooling operation with receptive field \( r_i \), performed \( R \) times for multi-scale fusion of peripheral background information. \( L^*_m \in \mathbb{R}^{D \times B} \) represents learnable textual prompts for category-agnostic objects, where \( B \) is the length of the fused background information. These tokens are concatenated with \( L^* \) to form a dual-branch structure for pixel-level target-background segmentation. Specifically, the \([cls]\) token is extracted from \( L^* \), and average pooling is applied to \( L^*_m \) to obtain global referring prototypes \( L_{\text{fg}} \) and \( L_{\text{bg}} \) for the target and background, respectively (\( L_{\text{fg}}, L_{\text{bg}} \in \mathbb{R}^D \)). These features are matched with \( V_{\text{pred}} \) to predict pixel-wise probabilities for the foreground and background regions, respectively, as follows:
\begin{equation}
\label{eq:fg_bg_projection}
\begin{cases}
O_{\text{fg}}^{i,j} = I_{\text{proj}}(V_{\text{pred}}) \circledcirc
 R_{\text{proj}}(L_{\text{fg}}), \\
O_{\text{bg}}^{i,j} = I_{\text{proj}}(V_{\text{pred}}) \circledcirc
 R_{\text{proj}}(L_{\text{bg}}).
\end{cases}
\end{equation}
The outputs \( O_{\text{fg}}, O_{\text{bg}} \in \mathbb{R}^{H_1 \times W_1} \), are interpolated to match the input image dimensions \( H \times W \) and compared with the corresponding ground truth to compute the cross-entropy loss:
\begin{equation}
\label{eq:cross_entropy_loss}
\begin{cases}
L_{\text{fg}} = -\sum_{i,j} G_{\text{fg}}^{i,j} \log(O_{\text{fg}}^{i,j}), \\
L_{\text{bg}} = -\sum_{i,j} G_{\text{bg}}^{i,j} \log(O_{\text{bg}}^{i,j}).
\end{cases}
\end{equation}
Where \( G_{\text{fg}}^{i,j} \) and \( G_{\text{bg}}^{i,j} \) represent the ground truth labels for the target and background classes at pixel \((i,j)\), respectively. The total cross-entropy loss \cite{zhang2018generalized} is computed as:
\begin{equation}
\label{eq:total_loss1}
L_{\text{ce}} = \lambda L_{\text{fg}} + (1-\lambda) L_{\text{bg}}
\end{equation}
Here, \( \lambda \) is the weight of the foreground-background prediction in the mask loss, involved in backpropagation during training.

\subsection{Dual-Modal Object Learning}
We introduce a dual-modality target learning strategy to address the challenge of capturing semantic information in remote sensing images. Initially, the cross-modal features \( V^* \) and \( L^* \), enriched with target-specific information from the preceding stage, are fed along with the masked semantic tokens \( L_m \). This approach aims to guide the reconstruction of object descriptions within \( L_m \). Specifically, the operation of the Joint Reconstruction (JR) module is implemented using a transformer decoder as follows:
\begin{equation}
\label{eq:lang_transform}
L_{re} = \text{CrossAtt}\left(L_m, \text{CrossAtt}\left(V^*, L^*\right)\right)
\end{equation}
Here, \(\text{CrossAtt}( \cdot )\) represents the cross-modal attention\cite{vaswani2017attention}. In this process, we first use \( V^* \) as the query and \( L^* \) as the key and value for text information enrichment and alignment. Then, \( L_m \) is used as the query to retrieve the missing information in the text from the fused multimodal features.
For the rephrased object description \( L_{re} \) and the original \( L_0 \), we define the reconstruction loss as follows:
\begin{equation}
\label{eq:reconstruction_loss}
L_{\text{re}} = \frac{1}{N} \sum_{i=1}^N \| L_{re}^{(i)} - \bar{L}_0^{(i)} \|^2
\end{equation}
Where \( N \) is the number of tokens, \( L_{re}^{(l)} \) and \( L_0^{(l)} \) represent the \( l \)-th token in the rephrased and original descriptions, respectively. Here, \( \bar{L}_0 \) indicates that the gradient of \( L_0 \) is stopped during this stage.
Finally, combining the reconstruction loss \( L_{\text{re}} \) with the cross-entropy loss \( L_{\text{ce}} \) from the previous section, we derive the final loss $L_{\text{total}}$ for backpropagation as:
\begin{equation}
L_{\text{total}} = L_{\text{ce}} + \eta \cdot L_{\text{re}}
\label{eq:total_loss}
\end{equation}
Here, \(\eta\) is a hyperparameter that balances the importance of the reconstruction loss \( L_{\text{re}} \) and the cross-entropy loss \( L_{\text{ce}} \). The total loss for the batch is computed as the sum of \( L_{\text{ce}} \) and \( L_{\text{re}} \).

\begin{table*}[htbp]
    \caption{Comparison with state-of-the-art methods on the proposed RefSegRS dataset. R-101 and Swin-B represent ResNet-101 and base Swin Transformer models, respectively. The best result is bold.}
    \label{tab:comparison_refsegrs}
    \centering
    \resizebox{\textwidth}{!}{%
    \begin{tabular}{c|c|c|c|c|c|c|c|c|c|c|c|c|c|c|c|c}
        \toprule
        \multirow{2}{*}{Method} & \makecell{Visual} & \makecell{Text} & \multicolumn{2}{c|}{P@0.5} & \multicolumn{2}{c|}{P@0.6} & \multicolumn{2}{c|}{P@0.7} & \multicolumn{2}{c|}{P@0.8} & \multicolumn{2}{c|}{P@0.9} & \multicolumn{2}{c|}{oIoU} & \multicolumn{2}{c}{mIoU} \\
        \cline{4-17} 
        & \makecell{Encoder} & \makecell{Encoder} & Val & Test & Val & Test & Val & Test & Val & Test & Val & Test & Val & Test & Val & Test \\
        \midrule
        BRINet \cite{hu2020bi} & R-101 & LSTM & 36.86 & 20.72 & 35.53 & 14.26 & 19.93 & 9.87 & 10.66 & 2.98 & 2.84 & 1.14 & 61.59 & 58.22 & 38.73 & 31.51 \\
        LSCM \cite{hui2020linguistic} & R-101 & LSTM & 56.82 & 31.54 & 41.24 & 20.41 & 21.85 & 9.51 & 12.11 & 5.29 & 2.51 & 0.84 & 62.82 & 61.27 & 40.59 & 35.54 \\
        CMPC \cite{huang2020referring} & R-101 & LSTM &  46.09&  32.36&  26.45&  14.14&  12.76&  6.55&  7.42&  1.76&  1.39&  0.22&  63.55&  55.39&  42.08&  40.63\\
        CMSA \cite{ye2019cross} & R-101 & None & 39.24 & 28.07& 38.44 & 20.25& 20.39 & 12.71& 11.79 & 5.61& 1.52 & 0.83& 65.84& 64.53& 43.62 & 41.47\\
        RRN \cite{li2018rrn} & R-101 & LSTM & 55.43 & 30.26 & 42.98 & 23.01 & 23.11 & 14.87 & 13.72 & 7.17 & 2.64 & 0.98 & 69.24 & 65.06 & 50.81 & 41.88 \\
        CMPC+ \cite{liu2021cross} & R-101 & LSTM &56.84&  49.19&  37.59&  28.31&  20.42&  15.31&  10.67&  8.12&  2.78&  2.55&  70.62&  66.53&  47.13&  43.65  \\
        CARIS \cite{liu2023caris} & Swin-B & BERT & 68.45 & 45.40 & 47.10 & 27.19 & 25.52 & 15.08 & 14.62 & 8.87 & 3.71 & 1.98 & 75.79 & 69.74 & 54.30 & 42.66 \\
        CRIS \cite{wang2022cris} & R-101 & CLIP & 53.13 & 35.77 & 36.19 & 24.11 & 24.36 & 14.36 & 11.83 & 6.38 & 2.55 & 1.21 & 72.14 & 65.87 & 53.74 & 43.26 \\
        RefSegformer \cite{wu2024towards} & Swin-B & BERT & 81.67 & 50.25 & 52.44 & 28.01 & 30.86 & 17.83 & 17.17 & 9.19 & 5.80 & 2.48 & 77.74 & 71.13 & 60.44 & 47.12 \\
        LAVT \cite{yang2022lavt} & Swin-B & BERT & 80.97 & 51.84 & 58.70 & 30.27 & 31.09 & 17.34 & 15.55 & 9.52 & 4.64 & 2.09 & 78.50 & 71.86 & 61.53 & 47.40 \\
        RIS-DMMI \cite{hu2023beyond} & Swin-B & BERT & 86.17 & 63.89 & 74.71 & 44.30 & 38.05 & 19.81 & 18.10 & 6.49 & 3.25 & 1.00 & 74.02 & 68.58 & 65.72 & 52.15 \\
        LGCE \cite{yuan2024rrsis} & Swin-B & BERT &  90.72&  73.75&  86.31&  61.14&  71.93&  39.46&  32.95&  16.02&  10.21&  5.45&  83.56&  76.81&  72.51&  59.96\\
        RMSIN \cite{liu2024rotated} & Swin-B & BERT &  93.97&  79.20&  89.33&  65.99&  74.25&  42.98&  29.70&  16.51&  7.89&  3.25&  82.41&  75.72&  73.84&  62.58\\
        \rowcolor{gray!20}\ours (Ours)& Swin-B & BERT &  \textbf{95.13} & \textbf{83.60}& \textbf{94.20} & \textbf{75.07}& \textbf{90.72} & \textbf{62.69}& \textbf{68.91} & \textbf{34.40}& \textbf{19.49} & \textbf{9.14}& \textbf{87.92} & \textbf{80.57}& \textbf{80.61} & \textbf{67.95}\\
        \bottomrule
    \end{tabular}%
    }
    \end{table*}

\begin{table*}[htbp]
    \caption{Comparison with state-of-the-art methods on the proposed RRSIS-D dataset. R-101 and Swin-B represent ResNet-101 and base Swin Transformer models, respectively. The best result is bold.}
    \label{tab:comparison}
    \centering
    \resizebox{\textwidth}{!}{%
    \begin{tabular}{c|c|c|c|c|c|c|c|c|c|c|c|c|c|c|c|c}
        \toprule
        \multirow{2}{*}{Method} & \makecell{Visual} & \makecell{Text} & \multicolumn{2}{c|}{Pr@0.5} & \multicolumn{2}{c|}{Pr@0.6} & \multicolumn{2}{c|}{Pr@0.7} & \multicolumn{2}{c|}{Pr@0.8} & \multicolumn{2}{c|}{Pr@0.9} & \multicolumn{2}{c|}{oIoU} & \multicolumn{2}{c}{mIoU} \\
        \cline{4-17} 
        & \makecell{Encoder} & \makecell{Encoder} & Val & Test & Val & Test & Val & Test & Val & Test & Val & Test & Val & Test & Val & Test \\
        \midrule
       RRN \cite{li2018rrn} & R-101 & LSTM & 51.09 & 51.07 & 42.47 & 42.11 & 33.04 & 32.74 & 20.80 & 21.57 & 6.14 & 6.37 & 66.53 & 66.43 & 46.06 & 45.64 \\
        CMSA \cite{ye2019cross} & R-101 & None & 55.68 & 55.32 & 48.04 & 46.45 & 38.27 & 37.43 & 26.55 & 25.39 & 9.02 & 8.15 & 69.68 & 69.39 & 48.85 & 48.54 \\
        LSCM \cite{hui2020linguistic} & R-101 & LSTM & 57.12 & 56.02 & 48.04 & 46.25 & 37.87 & 37.70 & 26.37 & 25.28 & 7.93 & 8.27 & 69.28 & 69.05 & 50.36 & 49.92 \\
        CMPC \cite{huang2020referring} & R-101 & LSTM & 57.93 & 55.83 & 48.85 & 47.40 & 38.50 & 36.94 & 25.28 & 25.45 & 9.31 & 9.19 & 70.15 & 69.22 & 50.41 & 49.24 \\
        BRINet \cite{hu2020bi} & R-101 & LSTM & 58.79 & 56.90 & 49.54 & 48.77 & 39.65 & 39.12 & 28.21 & 27.03 & 9.19 & 8.73 & 70.73 & 69.88 & 51.14 & 49.65 \\
        CMPC+ \cite{liu2021cross} & R-101 & LSTM & 59.19 & 57.65 & 49.36 & 47.51 & 38.67 & 36.97 & 25.91 & 24.33 & 8.16 & 7.78 & 70.14 & 68.64 & 51.41 & 50.24 \\
         CRIS \cite{wang2022cris} & R-101 & CLIP & 56.44 & 54.84 & 47.87 & 46.77 & 39.77 & 38.06 & 29.31 & 28.15 & 11.84 & 11.52 & 70.08 & 70.46 & 50.75 & 49.69 \\
        RefSegformer \cite{wu2024towards} & Swin-B & BERT & 64.22 & 66.59 & 58.72 & 59.58 & 50.00 & 49.93 & 35.78 & 33.78 & 24.31 & 23.30 & 76.39 & 77.40 & 58.92 & 58.99 \\
        LGCE \cite{yuan2024rrsis} & Swin-B & BERT & 68.10 & 67.65 & 60.52 & 61.53 & 52.24 & 51.45 & 42.24 & 39.62 & 23.85 & 23.33 & 76.68 & 76.34 & 60.16 & 59.37 \\
        RIS-DMMI \cite{hu2023beyond} & Swin-B & BERT & 70.40 & 68.74 & 63.05 & 60.96 & 54.14 & 50.33 & 41.95 & 38.38 & 23.85 & 21.63 & 77.01 & 76.20 & 60.72 & 60.12 \\
        LAVT \cite{yang2022lavt} & Swin-B & BERT & 69.54 & 69.52 & 63.51 & 63.61 & 53.16 & 53.29 & 43.97 & 41.60 & 24.25 & 24.94 & 77.59 & 77.19 & 61.46 & 61.04 \\
        CARIS \cite{liu2023caris} & Swin-B & BERT & 71.61 & 71.50 & 64.66 & 63.52 & 54.14 & 52.92 & 42.76 & 40.94 & 23.79 & 23.90 & 77.48 & 77.17 & 62.88 & 62.17 \\

        RMSIN \cite{liu2024rotated} & Swin-B & BERT & 74.66 & 74.26 & 68.22 & 67.25 & 57.41 & 55.93 & 45.29 & 42.55 & 24.43 & 24.53 & 78.27 & 77.79 & 65.10 & 64.20 \\
        \rowcolor{gray!20}\ours (Ours) & Swin-B & BERT & \textbf{77.87} & \textbf{75.93} & \textbf{71.26} & \textbf{69.92} & \textbf{60.11} & \textbf{59.29} & \textbf{47.53} & \textbf{46.25} & \textbf{27.70} & \textbf{27.46} & \textbf{79.29} & \textbf{79.23} & \textbf{66.89} & \textbf{66.04} \\
        \bottomrule
    \end{tabular}%
    \vspace{-6mm}
    }
    
\end{table*}

\section{EXPERIMENTS}
\subsection{Metrics and Datasets}
To evaluate the performance of our \ours model on the RRSIS task, we employ three key metrics: precision at different IoU thresholds (Pr@0.5 to Pr@0.9), mean Intersection-over-Union (mIoU), and overall Intersection-over-Union (oIoU). Precision at different thresholds (Pr) captures the segmentation accuracy at varying levels of strictness, from Pr@0.5 to Pr@0.9, providing insights into the model's ability to handle different segmentation challenges. mIoU, on the other hand, treats both small and large objects equally, ensuring a balanced evaluation of the model's performance across all object sizes. In contrast, oIoU focuses more on large-scale targets by computing the global IoU across all pixels, reflecting the overall segmentation performance irrespective of category distribution. Together, these metrics comprehensively assess the model's segmentation capabilities. 

In this study, we utilize two benchmark datasets, RefSegRS \cite{yuan2024rrsis} and RRSIS-D \cite{liu2024rotated}, designed explicitly for Referring Remote Sensing Image Segmentation (RRSIS) tasks. These datasets offer diverse and challenging scenarios, providing a comprehensive benchmark for evaluating the performance of models in understanding spatially complex and linguistically nuanced segmentation tasks. We conduct experiments on these two datasets to validate the effectiveness of our proposed \ours on the RRSIS task.

\subsubsection{RefSegRS}
The RefSegRS dataset consists of 4,420 image-expression-mask triplets, which are divided into 2,172 samples for training, 413 for validation, and 1,817 for testing. Each image has a resolution of 512$\times$512 pixels. The dataset spans a variety of object categories commonly found in remote sensing imagery, including buildings, vehicles, vegetation, and water bodies, annotated with fine-grained segmentation masks corresponding to natural language expressions.

\subsubsection{RRSIS-D}
The RRSIS-D dataset is an enormous collection comprising 17,402 triplets, 12,181 samples for training, 1,740 for validation, and 3,481 for testing. Each image in this dataset is of high resolution, with a size of 800$\times$800 pixels. This dataset includes a broader range of spatial scales and object orientations, making it suitable for evaluating models that handle multi-scale and rotational challenges. The object categories in RRSIS-D cover urban infrastructure, agricultural landscapes, and natural scenes, offering a rich diversity for segmentation tasks.
\renewcommand\arraystretch{1.0}
\setlength{\tabcolsep}{1.6mm}
\begin{table}[ht]
\centering
\caption{The results on each category of RefSegRS Dataset. The best performance is bold.}
\label{tab:clsresults1}
\begin{tabular}{@{}c|c|c|c@{}}
\toprule
\textbf{Category}                & \textbf{RMSIN} & \textbf{LGCE} & \textbf{Ours} \\ \midrule
Road                             & 75.52 & 77.78 & \textbf{82.36} \\ 
Vehicle                          & 63.66 & 63.92 & \textbf{70.79} \\ 
Car                              & 63.02 & 61.70 & \textbf{69.19} \\ 
Van                              & 52.81 & 52.86 & \textbf{57.47} \\ 
Building                         & 82.52 & 85.01 & \textbf{86.67} \\ 
Truck                            & 51.71 & 59.49 & \textbf{76.16} \\ 
Trailer                          & \textbf{42.60} & 41.27 & 37.34 \\ 
Bus                              & 41.43 & 46.74 & \textbf{49.81} \\ 
Road Marking                     & 18.19 & \textbf{18.46} & 9.49 \\ 
Bikeway                          & 53.25 & 49.84 & \textbf{64.76} \\ 
Sidewalk                         & 52.16 & 59.20 & \textbf{67.66} \\ 
Low Vegetation                   & 39.27 & \textbf{51.59} & 39.86 \\ 
Impervious Surface               & 79.16 & 81.41 & \textbf{85.62} \\ \midrule
\textbf{Average}                 & 55.02& 51.64& \textbf{61.32}\\ \bottomrule
\end{tabular}
\vspace{0mm}
\end{table}
\renewcommand\arraystretch{1.0}
\setlength{\tabcolsep}{1.6mm}
\begin{table}[ht]
\centering
\vspace{-2mm}
\caption{The results on each category of RRSIS-D Dataset. The best performance is bold.}
\label{tab:clsresults2}
\begin{tabular}{@{}c|c|c|c@{}}
\toprule
\textbf{Category}                & \textbf{RMSIN} & \textbf{LGCE} & \textbf{Ours} \\ \midrule
Airport                         & 53.90          & 53.71& \textbf{56.88}\\
Golf field                      & 80.51          & 77.41& \textbf{82.63}\\
Expressway service area         & 70.58          & 69.12& \textbf{78.38}\\
Baseball field                  & \textbf{90.49} & 85.98& 90.00\\ 
Stadium                         & 88.57 & 86.31& \textbf{89.20}\\ 
Ground track field              & 93.63          & 91.66& \textbf{93.93}\\ 
Storage tank                    & 85.93          & \textbf{86.19}& 85.88\\ 
Basketball court                & \textbf{79.13} & 68.62& 78.51\\ 
Chimney                         & 85.81          & 79.66& \textbf{87.25}\\ 
Tennis court                    & 76.32          & 67.00& \textbf{78.09}\\ 
Overpass                        & 77.62          & 73.81& \textbf{79.29}\\ 
Train station                   & \textbf{62.31} & 59.40& 61.71\\ 
Ship                            & 81.13 & 77.08& \textbf{81.22}\\ 
Expressway toll station         & \textbf{84.41} & 78.36& 80.43\\ 
Dam                             & 61.32          & 59.09& \textbf{63.11}\\ 
Harbor                          & 40.92          & 30.13& \textbf{41.63}\\
Bridge                          & 67.56          & 63.23& \textbf{67.87}\\
Vehicle                         & 61.45          & 61.41& \textbf{73.41}\\ 
Windmill                        & 62.55          & 59.74& \textbf{62.75}\\ \midrule
\textbf{Average}                & 73.90          & 69.89& \textbf{75.38}\\ \bottomrule
\end{tabular}
\end{table}

\subsection{Experimental Settings}
This paper implements the proposed method using PyTorch. Following\cite{yuan2024rrsis,liu2024rotated}, we adopt the Swin Transformer Base\cite{liu2021swin}, pre-trained on ImageNet-22K\cite{he2022partimagenet}, as the visual backbone to ensure robust feature extraction, and utilize the BERT-base model\cite{devlin2018bert} from the HuggingFace Transformers library\cite{wolf2020transformers} as the text encoder, comprising 12 layers with an embedding size of 768. The maximum token length for descriptive text is set to 20 to accommodate concise language inputs. To enable hierarchical cross-modal interaction, the 12 BERT layers are grouped into four stages of three layers each, aligning with the hierarchical visual features extracted by the Swin Transformer and facilitating effective feature fusion.

The loss function is a combination of \( L_{\text{fg}} \), \( L_{\text{bg}} \), and \( L_{\text{re}} \), where the hyperparameters \(\lambda\) and \(\eta\), representing the weights of \( L_{\text{bg}} \) and \( L_{\text{re}} \), are set to 0.6 and 0.1, respectively. 
For the datasets RefSegRS\cite{yuan2024rrsis} and RRSIS-D\cite{liu2024rotated}, the input image size is set to 512×512. During training, we employ the AdamW optimizer, setting the initial learning rates for the encoder and other modules to 0.00001 and 0.0001. Following a "poly" policy with a fixed power of 0.9, these rates are annealed to zero. The model is trained for 50 epochs on four NVIDIA 3090 GPUs with a batch size of 8. During the inference stage, the prediction result is determined by the higher value of final logits obtained from the Fg-Bg Predictor.
\begin{figure*}
    \centering
    \includegraphics[width=1\linewidth]{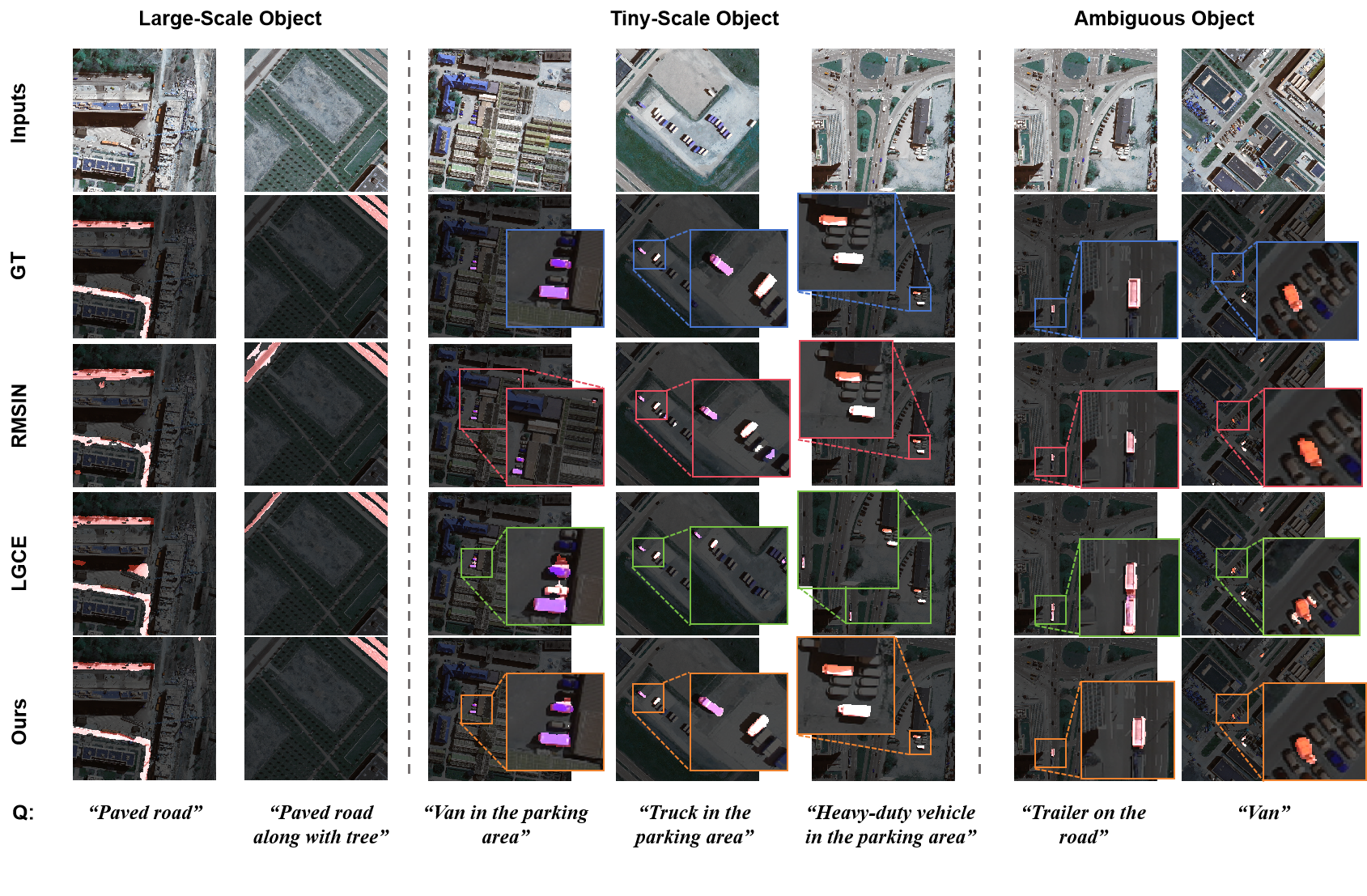}
    \caption{Qualitative comparisons between \ours and the previous SOTA methods on RefSegRS datasets.}
    \label{fig:visrefseg}
\end{figure*}
\begin{figure*}
    \centering
    \includegraphics[width=1\linewidth]{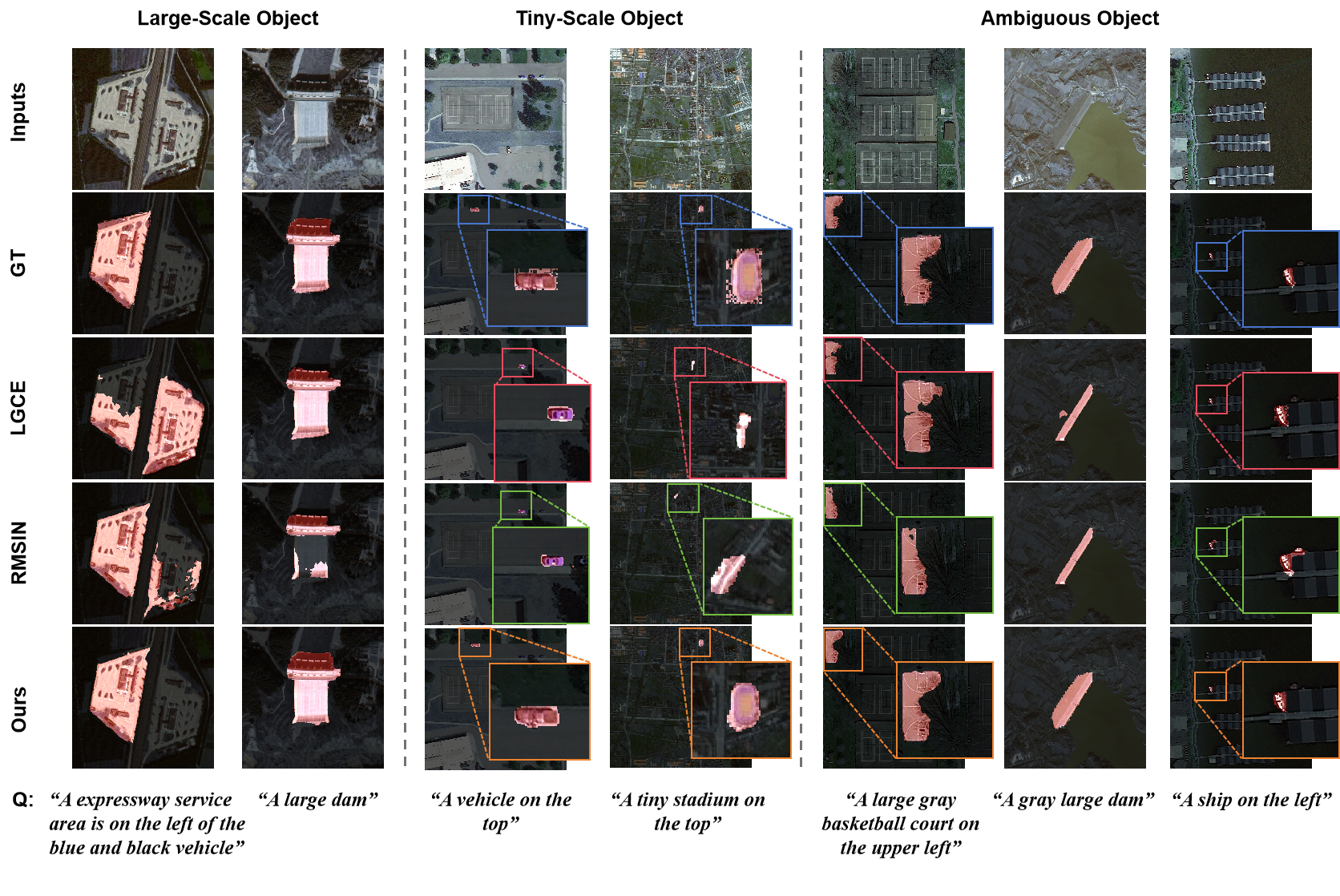}
    \caption{Qualitative comparisons between \ours and the previous SOTA methods on RRSIS-D datasets.}
    \label{fig:visrrsisd}
\end{figure*}
\begin{figure}
    \centering
    \includegraphics[width=0.9\linewidth]{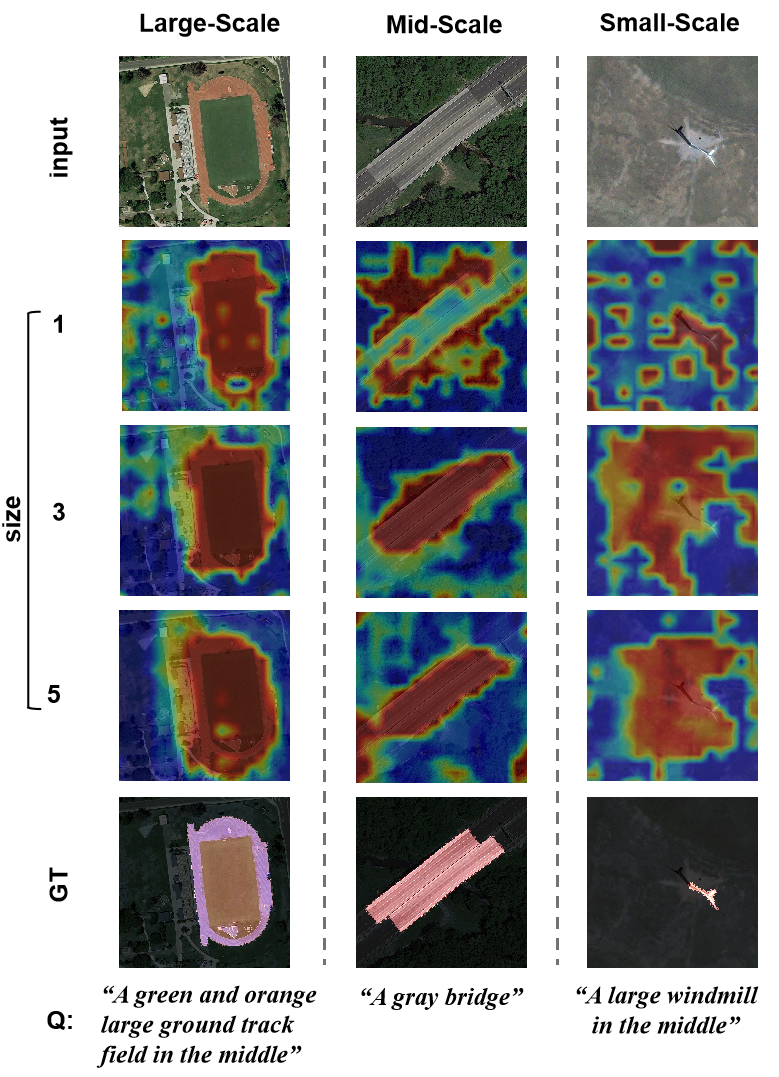}
    \caption{Visualization results for feature representations of different spatial correlation scales. Size 1, 3, 5 represent the unfold sizes used for visualization.}
    \label{fig:heatmap}
\end{figure}

\subsection{Performance Comparison}
Our method outperforms existing state-of-the-art approaches in the remote sensing referring segmentation task. Tables \ref{tab:comparison_refsegrs} and \ref{tab:comparison} present experiments on the RefSegRS and RRSIS-D datasets, comparing \ours with CNN-based methods like RRN \cite{li2018rrn}, CMPC \cite{huang2020referring}, BRINet \cite{hu2020bi}, and Transformer-based methods like DMMI \cite{hu2023beyond}, LAVT \cite{yang2022lavt}, and CARIS \cite{liu2023caris}. These methods, primarily designed for natural images, perform significantly worse on remote sensing data. We also compare with LGCE \cite{yuan2024rrsis} and RMSIN \cite{liu2024rotated}, which have achieved partial state-of-the-art results on RRSIS. Our approach consistently outperforms them, achieving state-of-the-art performance, especially in high-precision and fine-grained segmentation, with notable improvements in Pr@0.8, Pr@0.9, and mIoU.

\subsubsection{Quantitative Evaluations on RefSegRS} 

The targets in the RefSegRS dataset are characterized by scattered distributions, large shape variations, and a high proportion of small objects. Overall, \ours achieves state-of-the-art (SOTA) performance across all metrics, significantly surpassing LGCE, the second-best method. These substantial improvements confirm the effectiveness of \ours, which consistently outperforms LGCE on both the validation and test sets of the RefSegRS dataset.  

Specifically, \ours achieves a 7.99\% improvement in mIoU compared to LGCE, increasing from 59.96\% to 67.95\%, highlighting its strong capability to handle multi-scale objects. This improvement stems from the dual-directional spatial association design with multi-receptive fields, enhancing the model’s precision and responsiveness in spatial dimensions for individual instances. Moreover, the significant gains in Pr@0.6 to Pr@0.9, with improvements of 75.07\%, 62.69\%, and 34.40\% respectively, demonstrate that \ours excels in scenarios requiring finer-grained segmentation, achieving 10\% to 20\% higher performance compared to previous methods. This advantage, attributed to our Target-Background TwinStream Decoder (T-BTD), enables the model to better distinguish ambiguous objects and category-agnostic targets with blurred boundaries.  

To further verify the effects of our design, we calculated the average IoU for different object categories shown in Table \ref{tab:clsresults1}. For fine-grained small objects such as cars (69.19\%), vans (57.47\%), buses (49.81\%), and trucks (76.16\%), \ours delivers significantly superior results, outperforming other methods by approximately 8\% on average. Additionally, for ambiguous objects like various types of roads (82.36\%), buildings (86.67\%), bikeways (64.76\%), and sidewalks (67.66\%), \ours achieves notably high mIoU scores. This confirms the model’s exceptional entity binding and localization capabilities when dealing with small or shape-variant objects. 

More importantly, the average IoU for all categories with \ours is at least 6.3\% higher than other methods, achieving 61.32\% compared to 55.02\%, demonstrating its superior and comprehensive segmentation performance. However, for categories like Road Marking (9.49\%) and Low Vegetation (39.86\%), which are widely distributed and visually redundant, the model’s ability to cover extensive areas is somewhat limited. This limitation arises despite \ours's strong focus on individual instances in complex scenes and presents an opportunity for further improvement in future work.



\subsubsection{Quantitative Evaluations on RRSIS-D}
The RRSIS-D dataset encompasses a broader range of target categories, including urban, rural, port, and industrial environments, with higher spatial resolution and increased complexity. As shown in Table \ref{tab:comparison}, \ours achieves state-of-the-art performance across all evaluation metrics, demonstrating its adaptability and effectiveness in such diverse scenarios.

In terms of localization precision, \ours consistently outperforms RMSIN by approximately 3\% in Pr@0.5 to Pr@0.7, achieving 75.93\%, 69.92\%, and 59.29\%, indicating its robust ability to locate diverse objects even in complex and cluttered environments. This improvement highlights the contribution of our Bi-directional Spatial Correlation (BSC) module, which effectively handles the semantic redundancy caused by the high-resolution and rich object diversity of remote sensing imagery. For IoU metrics, \ours achieves gains of 1.44\% in oIoU (from 77.79\% to 79.23\%) and 1.84\% in mIoU (from 64.20\% to 66.04\%) on the test set, further underscoring its superior segmentation accuracy.

Table \ref{tab:clsresults2} further demonstrates the multi-scale generalization capabilities of \ours. It excels in segmenting a wide range of objects, including small objects such as vehicles (73.41\%), windmills (62.75\%), and ships (81.22\%), as well as large targets like stadiums (89.20\%), airports (56.88\%), and various fields (90.00\% and 93.93\%). Moreover, it outperforms other methods in distinguishing challenging targets, such as chimneys (from 85.81\% to 87.25\%), courts (from 76.32\% to 78.09\%) and overpasses (from 77.62\% to 79.29\%), which are often confused with background regions.
Additionally, the RRSIS-D dataset includes richer entity attributes and longer textual descriptions. Even under these challenging conditions, \ours maintains the highest average precision (form 73.90\% to 75.38\%), attributed to the Dual-Modal Object Learning Strategy (D-MOLS) that enhances text parsing through reconstruction learning.

In summary, \ours delivers robust and precise segmentation performance on RRSIS-D, good at handling diverse, high-resolution, and textually complex scenes, making it a highly effective solution for real-world remote sensing segmentation tasks.



\subsubsection{Qualitative Comparison}
As shown in Fig. \ref{fig:visrefseg} and \ref{fig:visrrsisd}, \ours demonstrates strong performance across various remote sensing scenarios and for different segmentation targets. Compared to previous state-of-the-art models, our \ours model exhibits precise semantic understanding and instance differentiation, particularly for small redundant objects and fine-grained targets in remote sensing images. For example, in the fourth and seventh columns, \ours can accurately identify vehicles with varying shapes and colors, effectively avoiding interference from similar categories. Furthermore, our model performs better in segmenting large objects and ambiguous targets, thanks to our \textbf{T-BTD} design, which allows the model to achieve superior localization for targets with low foreground-background distinction. 

Moreover, our model shows strong text-image alignment capabilities for objects with shallow semantic descriptions. As illustrated in Fig. \ref{fig:visrefseg}, the query \textit{'\textit{Van in the parking area}'} and its corresponding target, the lack of sufficient textual context often leads to misidentification in other models. In contrast, our model can effectively extract and utilize freely input textual information in various forms, highlighting the superiority of \ours in dual-modal analysis and alignment.

Furthermore, Fig. \ref{fig:heatmap} highlights the adaptability of our Bidirectional Spatial Correlation (BSC) module to multi-scale targets under varying receptive field scales. Large-scale targets like track fields achieve optimal feature representations at a spatial scale of 5, medium-sized targets such as bridges at a scale of 3, and small-scale targets like windmills at a scale of 1.

This demonstrates that our BSC module effectively captures hierarchical spatial correlations by leveraging multi-scale receptive fields, enabling precise alignment of image and text features across diverse target sizes. This adaptability is particularly valuable in remote sensing scenarios, where objects of varying scales and complex spatial distributions demand fine-grained segmentation and robust text-to-visual matching.

\subsection{Ablation Study}
We conducted a series of ablation experiments on the test subset of the RefSegRS dataset to verify the effectiveness of our method's core components. 
\begin{figure}
    \centering
    \includegraphics[width=1\linewidth]{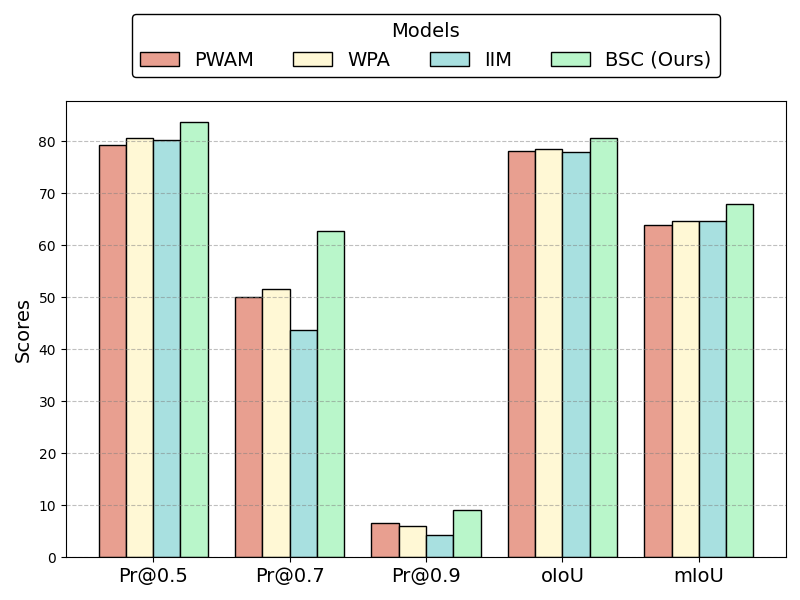}
    \vspace{-5mm}
    \caption{The comparisons of different cross-modal fusion mechanisms}
    \label{fig:fusion}
\end{figure}
\renewcommand\arraystretch{1.0}
\setlength{\tabcolsep}{1.8mm}
\begin{table}[htbp]
    \centering
    \caption{Ablation study comparing BSC and uni-directional V2L module at different stages.}
    \begin{tabular}{c|c|c|c|c|cc}
        \specialrule{.1em}{.05em}{.05em} 
        \multirow{2}{*}{\makecell[c]{Interaction Module}} & \multicolumn{4}{c|}{Stage} & \multicolumn{2}{c}{Metrics} \\
        \cline{2-7}
        & 1 & 2 & 3 & 4 & mIoU & oIoU \\
        \hline
        \multirow{4}{*}{\makecell[c]{BSC}} 
          & \checkmark &  &  &  & 62.67 & 78.99 \\
          & \checkmark & \checkmark &  &  & 65.55 & 79.47 \\
          & \checkmark & \checkmark & \checkmark &  & 65.90 & 79.60 \\
          & \cellcolor{gray!20}\checkmark & \cellcolor{gray!20}\checkmark & \cellcolor{gray!20}\checkmark & \cellcolor{gray!20}\checkmark & \cellcolor{gray!20}\textbf{67.95} & \cellcolor{gray!20}\textbf{80.57} \\
        \hline 
        \multirow{1}{*}{\makecell[c]{only V2L}} 
          & \checkmark & \checkmark & \checkmark & \checkmark & 67.13 & 79.70 \\
        \specialrule{.1em}{.05em}{.05em} 
    \end{tabular}
    \vspace{0mm}
    \label{tab:interaction_modules}
\end{table}
\subsubsection{Evaluation of Bidirectional Spatial Correlation}
Fig. \ref{fig:fusion} compares different cross-modal attention interaction mechanisms employed during feature extraction. Specifically, PWAM employs a unidirectional, multi-stage vision-to-text cross-attention mechanism, WPA utilizes parallel bidirectional attention learning, and IIM explores intra-scale information to enhance cross-modal interactions. In contrast, BSC, our proposed Bidirectional Spatial Correlation mechanism, surpasses these approaches with improvements of at least 2.04\% and 3.33\% in oIoU and mIoU, respectively. This demonstrates BSC's superior adaptability to remote sensing images and its enhanced capability for feature matching, especially for multi-scale targets and precise text localization.

Table \ref{tab:interaction_modules} further validates the effectiveness of bidirectional cross-modal interaction in feature extraction. Specifically, L2V refers to retaining only text-augmented visual feature matching while removing the visual-to-text pathway in BSC. Results indicate that the full BSC design significantly enhances semantic and visual alignment across depths and scales. Ablation studies on the multi-stage encoder reveal that bidirectional interactions—particularly in deeper layers—consistently outperform unidirectional approaches. Moreover, the BSC mechanism effectively leverages spatial information to align textual descriptions with targets of varying scales, as evidenced by its substantial improvements in mIoU.
\begin{figure}
    \centering
    \includegraphics[width=0.95\linewidth]{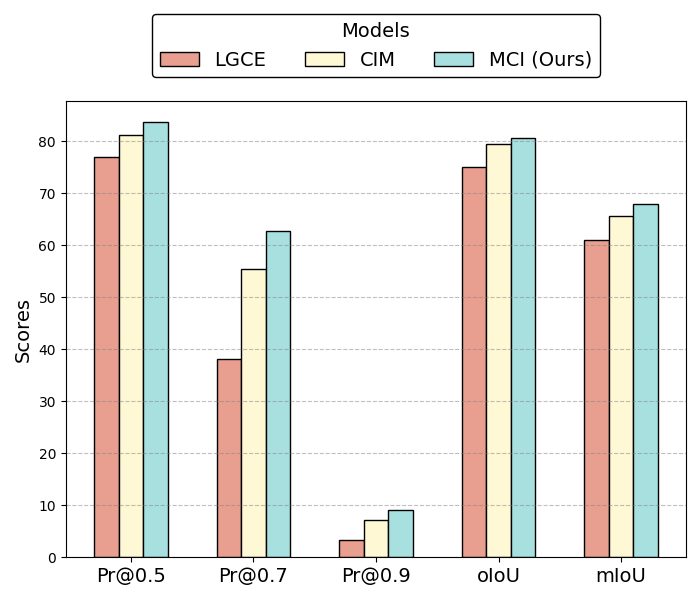}
    \vspace{-3mm}
    \caption{The comparisons of different multi-scale fusion mechanisms}
    \label{fig:scale}
\end{figure}
\subsubsection{Evaluation of Background Predictor}
\renewcommand\arraystretch{1.0}
\setlength{\tabcolsep}{1.6mm}
\begin{table}[htbp]
    \centering
    \footnotesize
    \vspace{-2mm}
    \caption{Ablation Study on the Impact of Background Token Length and Prior Knowledge in T-BTD.}
    \begin{tabular}{c|c|cc}
        \specialrule{0.1em}{0.05em}{0.05em} 
        \multirow{2}{*}{\makecell[c]{Number of \\ Bg Semantic Tokens}} & \multirow{2}{*}{Prior Knowledge} & \multicolumn{2}{c}{Metrics} \\
        & & mIoU & oIoU \\
        \hline
        3 & With & 66.80 & 80.30 \\
        \rowcolor{gray!20}5 & With & \textbf{67.95} & \textbf{80.57}  \\
        7 & With & 67.31 & 80.19 \\
        \hline
        5 & Without & 66.46 & 79.52 \\
        \specialrule{0.1em}{0.05em}{0.05em} 
    \end{tabular}
    \vspace{0mm}
    \label{tab:bi_attention_layers}
\end{table}
Fig. \ref{fig:scale} compares our proposed Multi-scale Context Integration (MCI) module with other multi-scale fusion mechanisms. LGCE uses a self-attention mechanism for parallel alignment across resolutions, while CIM employs cross-information interaction between multi-scale features at different stages. Our MCI module outperforms the second-best method by 1.11 and 2.41 in oIoU and mIoU, respectively, demonstrating superior multi-scale contextual integration. 

Table \ref{tab:bi_attention_layers} validates the effectiveness of background prediction in enhancing information richness and incorporating prior knowledge. Ablation studies show the best dual-stream performance when the number of Bg semantic tokens is set to 5, leveraging learnable parameters for semantic fitting of unknown classes. Additionally, using masked noun phrases as text references improves background prediction, as shown in the fourth row, with a ~0.7 drop in mIoU and oIoU observed when background prior knowledge is removed.
\renewcommand\arraystretch{1.0}
\setlength{\tabcolsep}{1.6mm}
\vspace{0mm} 
\begin{table}[htbp]
    \centering
    \footnotesize
    \vspace{-2mm}
    \caption{Ablation study on the effect of the T-BTD and D-MOLS.}
    \begin{tabular}{c|c|c|cc}
        \specialrule{.1em}{.05em}{.05em} 
        \multirow{2}{*}{\makecell[c]{\(L_{\text{fg}}\)}} & \multirow{2}{*}{\makecell[c]{\(L_{\text{bg}}\)}} & \multirow{2}{*}{\makecell[c]{\(L_{\text{re}}\)}} & \multicolumn{2}{c}{Metrics} \\
        &  &  & mIoU & oIoU \\
        \hline
        
        \usym{2714} & \usym{2718} & \usym{2718} & 64.92 & 75.13 \\
        \usym{2714} & \usym{2714} & \usym{2718} & 66.22 & 79.80 \\
        \usym{2714} & \usym{2718} & \usym{2714} & 65.09 & 76.26 \\
        \rowcolor{gray!20}\usym{2714} & \usym{2714} & \usym{2714} & \textbf{67.95} & \textbf{80.57} \\
        \specialrule{.1em}{.05em}{.05em} 
    \end{tabular}
    \label{tab:ablation_modules}
\end{table}

\subsubsection{Evaluation of Dual-Modal and Target-Background Learning Strategy}
Table \ref{tab:ablation_modules} validates the importance of the various losses designed in our model. When background prediction and text rephrasing are excluded from the training process, meaning their corresponding loss functions are not backpropagated, the model experiences a notable performance degradation. This decline becomes particularly evident when predictions for unknown classes are omitted, as demonstrated in rows 1, 3, and 4 of the table. Additionally, when the learning process of the semantic modality is neglected, the model's performance in terms of mIoU suffers considerably. This demonstrates that text reconstruction make the model to focus on learning and matching complex edge environments and high-fidelity segmentation, while the reconstruction process achieves a more accurate text-image alignment, which is better suited for remote sensing tasks.
 
\section{Conclusion}
In conclusion, we have presented a novel approach for the Referring Remote Sensing  Image Segmentation (RRSIS) task, named \ours. Our method effectively addresses the challenges of remote sensing imagery, particularly the vision-language gap and complex object segmentation. First, to bridge the substantial vision-language gap, we introduce the \textbf{bidirectional spatial correlation module}, enabling bidirectional multimodal feature interaction, thereby improving the alignment between visual features and textual descriptions in remote sensing tasks. To tackle the challenge of diverse object categories and precise localization, especially for small objects, we employ a \textbf{dual-modal object learning strategy}. This approach enhances fine-grained segmentation and improves the model's ability to handle objects at varying scales. Lastly, to address boundary delineation and ambiguous targets, we incorporate a \textbf{target-background twin-stream decoder}, which effectively handles discrete and blurred target boundaries by performing foreground-background joint prediction.These innovations collectively advance segmentation accuracy and robustness in the context of RRSIS. Extensive experiments on the RefSegRS and RRSIS-D datasets validate the effectiveness of our proposed modules and strategies. Our work contributes to advancing the remote sensing image segmentation field, providing a robust solution to referring segmentation in challenging environments.

\textbf{Limitations and Future Works:} Despite achieving remarkable results on the RRSIS task, our method has certain limitations. One potential issue is that our language backbone, BERT, may struggle with handling complex, domain-specific descriptions in remote sensing. With the rise of large language models (LLMs) \cite{achiam2023gpt, lai2024lisa}, integrating an LLM could help bridge the vision-language gap in remote sensing tasks. Another limitation is that while our method delivers precise segmentation, it cannot determine whether the described target area actually exists within the given image. This is due to the dataset bias, where each remote sensing image-language pair typically contains only one target. Future work should focus on developing more robust RRSIS methods to address these challenges and enable practical applications in real-world scenarios.
 
\newpage
\bibliographystyle{IEEEtran}
\bibliography{references_new}

\vfill

\end{document}